\documentclass{acm_proc_10ptArticle-sp}

\usepackage{xspace}
\usepackage{epsfig}
\usepackage{floatflt}
\usepackage{url}
\usepackage{amsmath}
\usepackage{ifthen}

\newcommand{\dolconcept}[1]{\texttt{#1}}

\newcommand{\individual}[1]{\texttt{#1}}

\newcommand{\fconcept}[1]{\texttt{#1}}

\newcommand{\domainconcept}[1]{\texttt{#1}}

\newcommand{\DOLCE}{DOLCE\xspace}

\newcommand{\DULextended}{DOLCE+DnS Ultralite\xspace}

\newcommand{\event}{event\xspace}
\newcommand{\events}{events\xspace}
\newcommand{\object}{object\xspace}
\newcommand{\objects}{objects\xspace}

\newcommand{\Objects}{Objects\xspace}

\begin{document}

\conferenceinfo{K-CAP'09,} {September 1--4, 2009, Redondo Beach, California, USA.} 
\CopyrightYear{2009}
\crdata{978-1-60558-658-8/09/09} 

\title{F---A Model of Events based on the\\ Foundational Ontology \DULextended}

\numberofauthors{1}
\author{
				 \alignauthor Ansgar Scherp, Thomas Franz, Carsten Saathoff, and Steffen Staab\\
	       \affaddr{University of Koblenz-Landau, Germany}\\
	       \email{\{scherp,franz,saathoff,staab\}@uni-koblenz.de}
	}

\date{\today}
\maketitle 

\begin{abstract}
The lack of a formal model of events hinders interoperability in distributed event-based systems.
In this paper, we present a formal model of events, called Event-Model-F. 
The model is based on the foundational ontology \DULextended (DUL) and provides comprehensive support to represent time and space, objects and persons, as well as mereological, causal, and correlative relationships between events.
In addition, the Event-Model-F provides a flexible means for event composition, modeling event causality and event correlation, and representing different interpretations of the same event.
The Event-Model-F is developed following the pattern-oriented approach of DUL, is modularized in different ontologies, and can be easily extended by domain specific ontologies.
 
\textbf{Note}: 
The Event-Model-F has been created in OWL and axiomatized in Description Logics. 
The ontologies have been republished in \url{https://github.com/ascherp/ontologies}
\end{abstract}


\section{Introduction}
\label{sec:intro}

\thispagestyle{empty}

The explicit modeling of events and event-based systems are increasingly gaining widespread attention by research and industry due to a couple of reasons.
Firstly, we find an increasing number of systems that are treating events, e.g., media delivery, surveillance video, or management of emergency incidents.
Secondly, a fastly growing number of intelligence-collecting devices such as sensors, CCTV, upload facilities, and others lead to an ubiquity of events being recognized and communicated. 
Thirdly, event detection, clustering, and annotation is and will be realized in many different software components and proprietary solutions using a large variety of internal data models.
Thus, multiple systems are connected for managing events resulting in a complex, so-called distributed event-based system~\cite{MuhlEtAlDEBS2006}. 
Such a distributed event-based system is a software system consisting of several components that are characterized by taking events as input and providing events as output. 
Existing solutions typically focus on processing low-level signals and actions, i.e., technical events that happen \textit{within} computerized systems~\cite{MuhlEtAlDEBS2006} like
~\cite{OberleMiddleware2006,EricssonBerndtssonREX2007,Zdonik2003,Vaculin2007,May2005}.

Our understanding of events is different from the technical, low-level events above.
We apply events to capture and represent human experience, i.e., to describe on a high-level the occurrences in which humans participate.
These events are subject to discussions and interpretations by humans.
They may be very complex and a variety of aspects need to be considered such as time and space, objects and persons involved, as well as mereological, causal, and correlative relationships between events.
Existing models for events like~\cite{WangEtAl2007,RaimondAbdallahEventOntology2007,IPTCEventML2008,DoerrCIDOC2007,Mueller2008,FrancoisEtAlVERL2005,WestermannJainE2007}
do not follow a systematic development approach.
They are conceptually narrow and their semantics is typically ambiguous.
This hinders interoperability of the different event-based components and event-based systems aggravating the treatment of events in already complex, distributed infrastructures.

In this paper, we present a formal representation of \events that allows for capturing and representing human experience, called Event-Model-F.
This representation allows easy interchange of event information between different event-based components and systems. 
Modeling such events is an interesting and challenging task as it involves among others 
causal 
relationships between \events and interpretations of the same event by different humans.
Such an event model is important in a large variety of domains such as emergency response, sports, news, law, and others.
As events and objects are essential for this paper and their concepts are of metaphysical interest, we introduce them in the next section.
A concrete scenario in Section~\ref{sec:scenario} motivates the need for a formal model on events. 
Section~\ref{sec:requirements} presents the requirements on the event model derived from an analysis of existing models. 
In Section~\ref{sec:f}, we describe the development of our Event-Model-F.
Our event model is based on the Descriptive Ontology for Linguistic and Cognitive Engineering (DOLCE)~\cite{GangemiEtAlDOLCE2002,DOLCE2003}.
DOLCE already has proved to be a good modeling basis for core ontologies such as~\cite{ArndtEtAlCOMM2007,FranzStaabArndtXCOSIM2007,OberleEtAlOntologiesForLargeSoftwareSystems2006,OberleEtAlSWIntO2007} and provides a formal modeling basis.
We aligned the Event-Model-F with the \DULextended (DUL) ontology\footnote{\url{http://ontologydesignpatterns.org/wiki/Ontology:DOLCE+DnS_Ultralite}}.
For designing the Event-Model-F, we have followed the pattern-oriented ontology design approach of DOLCE and DUL, respectively.
This approach provides native support for modularization of F and extension by domain specific ontologies. 
The use of our event model is demonstrated in Section~\ref{sec:use}. 
In Section~\ref{sec:relatedwork}, we present the extensive analysis of existing event models and compare our approach with related work in the area of low-level processing of technical events, before we conclude the paper.

\section{Events and Objects}
\label{sec:events}

In philosophical literature, there are different discussions of how to discriminate events from other categories~\cite{StanfordEncyclopediaEvents}.
One of them are objects.
Although not undisputed, there are standard differences between events and (physical) objects~\cite{StanfordEncyclopediaEvents}:
Events are said to \textbf{occur} or \textbf{happen}.
They are considered perduring entities that unfold over time, i.e., they take up time.
In contrast, material objects such as stones and chairs are said to \textbf{exist}. 
Such enduring entities unfold over space, i.e., they are in time.
As said, this metaphysical distinction is not uncontroversial as some philosophers consider objects as four-dimensional entities that extend across time just as they do across space~\cite{StanfordEncyclopediaEvents}. 
For our Event-Model-F, we follow \DULextended's design decision and distinguish \events from \objects. 
By this, we can be precise about the relationships that can occur between \events and \objects~\cite{OberleMiddleware2006}.

\section{Scenario}\label{sec:scenario}

\begin{figure}[htp]
\centering
\includegraphics[width=\columnwidth]{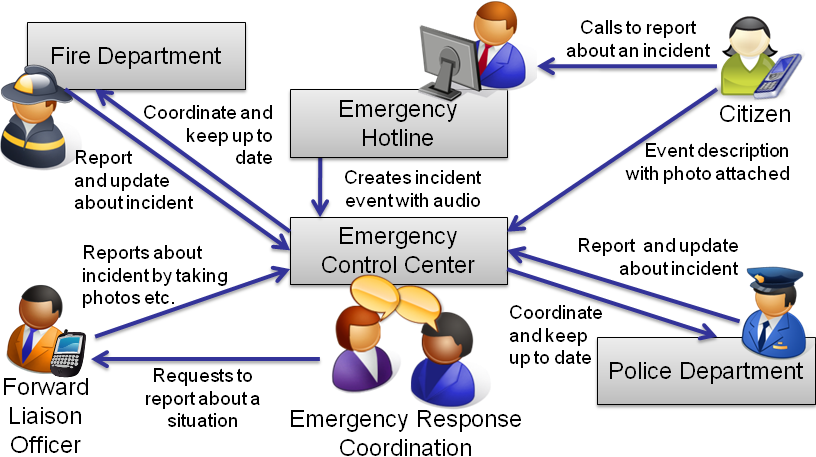}
\caption{A distributed event-based system for emergency response}
\label{fig:motivation}
\end{figure}

An example of a distributed event-based system is the emergency response use case of the EU project WeKnowIt\footnote{\url{http://www.weknowit.eu/}} depicted in Figure~\ref{fig:motivation}. 
Here, different professional entities are involved such as the emergency hotline, police department, fire department, emergency control center, and forward liaison officers. 
Please note again, the events we are modeling aim at capturing and representing human experience and thus are different from technical events as discussed in Section~\ref{sec:intro}.
The different aspects of events are marked with (\textit{$\langle$number$\rangle$}).

\newcommand{\Fmere}{i}
\newcommand{\Fpart}{ii}
\newcommand{\Fdocu}{iii}
\newcommand{\Fspac}{iv}
\newcommand{\Fcaus}{v}
\newcommand{\Fcorr}{vi}
\newcommand{\Ftime}{vii}
\newcommand{\Finte}{viii}

In an incident of a heavy storm a major flooding may happen.
(\Fmere)~During the flooding a power outage occurs. 
(\Fpart)~Some citizens are lacking power supply and are calling the emergency hotline to report about the outage. 
The officers at the emergency hotline record these calls and type in an event description for each call to document them in their system. 
(\Fdocu)~These events are annotated with information about the call and its recording and are automatically transferred to the system of the emergency control center.
In the course of the flood, (\Fmere) further events happen such as pumping out flooded cellars or rescuing people from their flooded homes.
(\Fspac)~Forward liaison officers drive to the different locations of the events to take photos and report them. 
Here, the event description is created using an application on their cell phone.
(\Fdocu)~A documentary support for the event is provided by attaching a photo to it and tagging it.
The event description is also send to the system of the emergency control center.
(\Fcaus)~Although the events of a flooded cellar and rescuing people from their homes might occur at different times and be located farer away from each other, they are probably both caused by the same flooding event.
(\Fcorr)~Thus, the events that happen during an incident correlate to each other.
These correlations are important to recognize to gain a full understanding of the emergency situation.
(\Ftime)~The emergency control center also receives event descriptions from the systems of the police department and fire department that happen during the incident.
(\Fcaus)~Based on the evidence of the event descriptions, the officers in the emergency control center use their system to formulate hypothetical events that might have caused the power outage.
(\Finte)~The officers conclude that there are two possible interpretations that might have caused the power outage, namely a snapped power pole close to the river or a problem with the power plant.

As depicted in Figure~\ref{fig:motivation}, several entities are involved in an emergency response using different event-based systems. 
These systems need to be connected through a common understanding of events in order to efficiently communicate between the emergency response entities. 
Thus, a formal representation of events is useful for the interoperability among the various computer systems as it provides machine accessible semantics.

\section{Requirements on F}
\label{sec:requirements}

We have derived functional as well as non-functional requirements on our Event-Model-F.
These base on an analysis of existing event models and related work in Section~\ref{sec:relatedwork}, the emergency response scenario in Section~\ref{sec:events}, and reported and own experience in designing core ontologies~\cite{OberleEtAlSWIntO2007,OberleEtAlOntologiesForLargeSoftwareSystems2006,ArndtEtAlCOMM2007,FranzStaabArndtXCOSIM2007}.

\subsection{Functional Requirements}
\label{sec:func-requirements}

To derive the functional requirements we have analyzed existing models in various domains such as music~\cite{RaimondAbdallahEventOntology2007}, journalism~\cite{WangEtAl2007}, multimedia~\cite{EkinEtAl2004,FrancoisEtAlVERL2005,WestermannJainE2007}, news~\cite{IPTCEventML2008}, cultural heritage~\cite{DoerrCIDOC2007}, and knowledge representation~\cite{Mueller2008}. 
The most comprehensive list of functional requirements are the six aspects defined for the event model E~\cite{WestermannJainE2007} 
and the journalism interrogatives of the Eventory system~\cite{WangEtAl2007}.
We have blend the aspects in E and interrogatives of the Eventory system and have synthesized them into our requirements.
For each requirement, we also explicitly refer to the scenario in Section~\ref{sec:scenario}. 

\textbf{(1) Participation of \objects in \events.}
Representing participation of living and non-living \objects such as people, animals, and other material objects in \events and the roles they play in \events.
In the scenario, an example of objects participating in events and different object roles are to be found with (\Fpart).

\textbf{(2) Temporal duration of \events} and \textbf{(3) spatial extension of \objects.}
As \events unfold over time (see Section~\ref{sec:events}), their temporal duration needs to be modeled.
This can be conducted using absolute or relative representations of points in time.
\Objects unfold over space.
Thus, modeling their spatial extension needs to be supported.
This can be also modeled using absolute or relative positioning.
In the emergency response scenario, events occur and objects exist at different points in time (\Ftime) and at different locations (\Fspac).

\textbf{(4) Structural relationships between events.}
We consider three kinds of structural relationships between events, namely \textbf{(4a) mereological}, \textbf{(4b) causal}, and \textbf{(4c) correlation relationships}.
The mereological relationship should be supported as \events are usually made up of other \events~\cite{QuintonObjectsAndEvents1979}. 
Causal relationships require the modeling of causes and effects and should support the integration and use of different causal theories as discussed, e.g., in~\cite{ItkonenCausality1983}.
Correlation refers to two events that have a common cause (cf.~\cite{ShipleyCausAndCorrelation2002}). 
It should be supported as it is typically easy to observe, while causality is very difficult to discover and, hence, often unknown.
The 
scenario illustrates the requirement of mereology (\Fmere), causality (\Fcaus), and correlation (\Fcorr). 

\textbf{(5) Documentary support for \events and \objects.} 
This requirement comprises the annotation of \events and their participating \objects with arbitrary information such as sensor data and media data.
It provides documentary support that a particular \event happened or \object exists.
In the scenario, the documentary support is indicated with (\Fdocu).

\textbf{(6) Event interpretations.}
Relations between events such as causality and correlation can be matter of subjectivity and interpretation.
For example, in a law-suit the parties involved may each claim that the other one is at fault.
Thus, the event model should support such different interpretations of the same event, i.e., provide different contextual points of view onto the same occurrences in the real world.
In the scenario, event interpretation is indicated with (\Finte).

\subsection{Non-functional Requirements}
\label{sec:nonfunc-requirements}

Above, we enumerated \emph{what} needs to be expressed by a common model of events. 
We now elaborate \emph{how} such a model needs to be designed in order to be applicable in scenarios such as the one illustrated in Section~\ref{sec:scenario}. The non-functional requirements are derived from reported and own experience in knowledge-based systems and knowledge representation~\cite{OberleEtAlSWIntO2007,OberleEtAlOntologiesForLargeSoftwareSystems2006,ArndtEtAlCOMM2007,FranzStaabArndtXCOSIM2007}.

\textbf{(a) Extensibility.}
Systems evolve over time, being extended, combined, and integrated. 
A core model for knowledge representation needs to support system evolution by being extensible towards new developments and functional requirements that arise. 
With respect to the Event-Model-F, extensibility means to be able to, e.g., include future aspects for describing events. 

\textbf{(b) Axiomatization and formal precision.}
The goal of a core model is to establish a common understanding in a particular domain in order to ensure interoperability through machine accessible semantics.
Here, a mechanism is required to exchange structured knowledge about events and objects between different event-based systems.
These event-based systems need to be able to automatically check the validity of the exchanged knowledge, not only with respect to its syntax, but more importantly with respect to the semantics. 
Thus, a core model like the Event-Model-F needs sufficient formality and axiomatization 
so that systems can reason about the represented knowledge and carry out semantic checks on its validity.

\textbf{(c) Modularity.}
While a core model needs to capture different structural knowledge, systems will commonly use only portions of it.
A modular design allows for selecting the parts of the model used.
It decreases complexity of the system in use and its development time and simplifies the acquisition of the model.

\textbf{(d) Reuseability.}
Different event-based systems may be built for different tasks and users in different domains. 
Thus, they will focus on different aspects of events. 
Based on the modularity requirement, a core model needs to support reuse of its modules despite of the different points of view on events that may be imposed by different domains while still guaranteeing formal precision of the overall knowledge base.
This is of particular benefit for the functional requirement of providing different interpretations of the same event as the common parts of the event descriptions can be reused.
In addition, reuseability is also seen with respect to existing domain knowledge.
Here, a common model for events shall be able to incorporate existing domain ontologies and make use of that domain knowledge rather than requiring to remodel it.

\textbf{(e) Separation of concerns.}
A core model needs to be applicable for arbitrary application domains.
It should also enable integration and reuse of models of domain-specific knowledge such as in emergency response or sports.
To this end, the domain-independent knowledge in the core model needs to be clearly separated from the domain-specific knowledge.
Thus, the Event-Model-F needs to provide a clear separation of the structural knowledge about events and objects from the domain knowledge.

\section{Design of the Event-Model-F}

\label{sec:f}

For designing the Event-Model-F and implementing the functional requirements, 
we have carefully aligned it with the \DULextended (DUL) ontology.
DUL defines the class \dolconcept{DUL:Event} next to the disjoint upper classes \dolconcept{DUL:Object}, \dolconcept{DUL:Abstract}, and \dolconcept{DUL:Quality}.  
The definition of \dolconcept{Event} has been specialized from the formal definition in \DOLCE as an entity that exists in time (cf.~discussion of \events and \objects in Section~\ref{sec:events}). 
The class \dolconcept{Object} stands for entities that exist in space such as living things as well as non-living and abstract things like social and cognitive entities. 
A \dolconcept{Quality} is a characteristic of an object or an event.
It has a value that is represented as a point or area in some \dolconcept{Abstract}.
The class \dolconcept{Abstract} represents value spaces, e.g., the space of natural numbers or the time of a day.  
In the Event-Model-F, we do not prescribe specific \dolconcept{Abstract}s that are to be used.
We rather refer to the generic \dolconcept{Abstract}s already defined in DUL such as the regions \dolconcept{DUL:TimeInterval}, \dolconcept{DUL:SpatioTemporalRegion}, and \dolconcept{DUL:SpaceRegion}.  

The functional requirements on our Event-Model-F are represented by specialized instantiations of the descriptions and situations (DnS) ontology pattern that is part of \DULextended.
We use DnS as it provides formally precise representations of different, contextualized views on events~\cite{GangemiDnS2008}. 
Thus, with DnS one can reify events and describe the n-ary relation that exists between multiple individuals of events and objects.
We use the DnS pattern as the representation of occurrences in the real world (i.e., the \events and \objects we are modeling) are subject to discussion and interpretation and may not be objectively observable.
The DnS pattern allows for representing different opinions about \events and their participating \objects.
This feature is not provided by the DOLCE participation relation. 

In the following, we introduce the ontology patterns of the Event-Model-F and illustrate them in diagrams.
With respect to the functional requirements in Section~\ref{sec:func-requirements}, the participation of \objects in \events~(1) is implemented by the participation pattern. 
It also provides for modeling the absolute time and location of \events~(2) and \objects~(3).
The mereology pattern, causality pattern, and correlation pattern implement the structural relationships between events~(4a-4c).
In addition, the mereology pattern allows for modeling the relative temporal relations and relative spatial relations between events~(2) and objects~(3).
The documentation pattern provides for annotating events~(5) and the interpretation pattern supports different event interpretations~(6).
Classes defined by the Event-Model-F are highlighted in the diagrams to show the alignment with classes of DUL. 
The axiomatization of the Event-Model-F has been conducted in Description Logics and is available online at: \url{http://isweb.uni-koblenz.de/eventmodel}.

\subsection{Participation Pattern}

\label{sec:participationpattern}

\begin{figure*}[htb]
 \centering
\includegraphics[width=\textwidth]{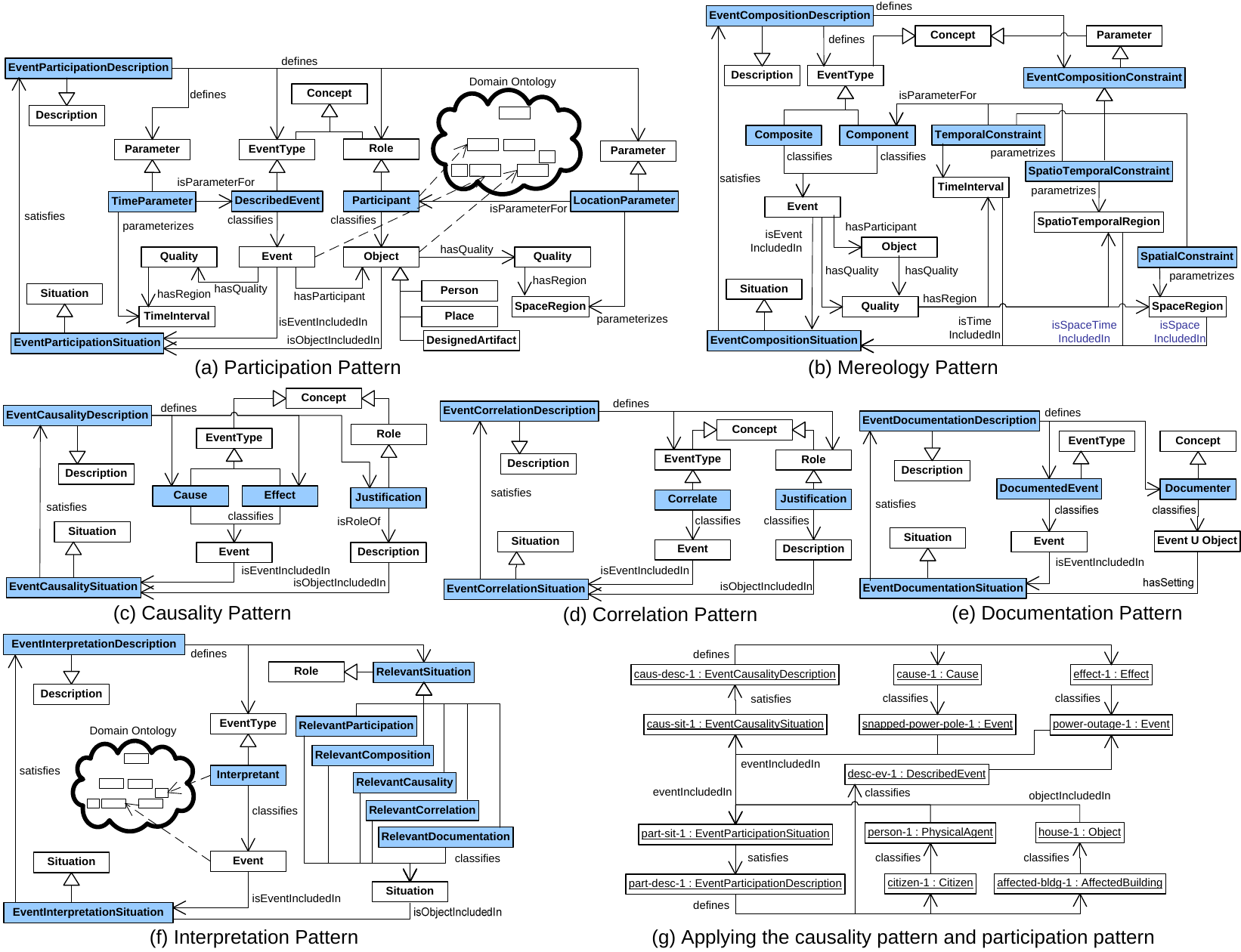}
 \caption{The patterns of F, namely (a)~participation, (b)~mereology, (c)~causality, (d)~correlation, (e)~documentation, and (f)~interpretation and (g)~example of applying the F ontology} 
 \label{fig:allpatterns}
 \label{fig:participation-pattern}
 \label{fig:decomposition-pattern}
 \label{fig:causality-pattern}
 \label{fig:correlation}
 \label{fig:documentation-pattern}
 \label{fig:interpretation-pattern}
 \label{fig:example}
\end{figure*}

The participation pattern enables to formally express the participation of \objects in \events. 
As shown in Figure~\ref{fig:participation-pattern}(a), participation is expressed by an \dolconcept{F:EventPar\-ticipationSituation} that \dolconcept{satisfies} an \fconcept{F:EventPar\-ticipationDescription}. 
The situation includes the \dolconcept{Event} being described and the \dolconcept{Object}s participating in this event. 
The \fconcept{EventParticipationDescription} classifies the described event and its participants by using the concepts \fconcept{F:DescribedEvent} (specialized from \dolconcept{DUL:EventType}) and the object role \fconcept{F:Participant} (specialized from \dolconcept{DUL:Role}).
The concept \dolconcept{Described\-Event} \dolconcept{classifies} the \dolconcept{Event} that is described by the participation pattern, e.g., the event of a flooded cellar.
Likewise, instances of \dolconcept{Participant} classify \objects as participants of the event.
For example, the citizen calling the emergency hotline to report about the flooded cellar. 
Instances of \dolconcept{Participant} can be roles defined in some domain ontology as indicated in Figure~\ref{fig:participation-pattern}(a). 
For example, an emergency response ontology may define the role of a person being affected, i.e., the emergency subject, and the role describing the rescue staff such as firemen.
Besides the role an object can play in a specific participation pattern, also the described event and its participating objects themselves can be defined in some domain ontology as indicated in Figure~\ref{fig:participation-pattern}(a).

The parameter \fconcept{F:LocationParameter} de\-scri\-bes the general spatial region where the objects are located.
It \dolconcept{DUL:parametrizes} a \dolconcept{SpaceRegion} and defines a property \dolconcept{DUL:\-is\-Parameter\-For} to the \fconcept{Par\-tic\-i\-pant} role. 
The \dolconcept{Object} that is classified by the \fconcept{Participant} has a \dolconcept{Quality} with the property \dolconcept{DUL:hasRegion} of a \dolconcept{Space\-Region}. 
Thus, using the \fconcept{F:LocationParameter} we can define the location(s) represented by \dolconcept{SpaceRegion}s that are relevant for describing the event in a given context.
For example, when quenching a house fire all firemen have their specific location within and around the building.
The \fconcept{LocationParameter} can then be used to describe in general that the firemen where at that specific house, e.g., in form of some longitude-latitude rectangular. 
Thus, we do not need to explicitly state 
where the individual firemen are.
The \fconcept{F:TimeParameter} describes the general temporal region when the event happened.
It \dolconcept{parametrizes} a \dolconcept{TimeInterval} and defines a property \dolconcept{is\-Parameter\-For} to the \fconcept{DescribedEvent} role.
For example, one can state that the house fire happened on June 13, 2006.

\subsection{Mereology Pattern}\label{sec:composition}

Events are commonly considered at different abstraction levels depending on the view and the knowledge of a spectator. 
For instance, the event of a flooded cellar may be considered as such or as part of the larger event of a flooding in which many other (smaller) incidents occur. 
The mereology pattern shown in Figure~\ref{fig:decomposition-pattern}(b) enables expressing such mereological relations as composition of events. 
The composite event is the ``whole'' and the component events are its ``parts''.
Formally, a \dolconcept{F:EventCompositionSituation} includes one instance of an \event that is classified by the concept \dolconcept{F:Composite} and many events classified as its \dolconcept{F:Com\-ponent}(s).
Accordingly, an \dolconcept{EventCompositionSitu\-ation} \dolconcept{satisfies} a \dolconcept{F:CompositionDescription} that \dolconcept{defines} the concepts \dolconcept{Composite} and \dolconcept{Component} for classifying the composite event and its component events.

Events that play the \fconcept{Component} role may be further qualified by temporal, spatial, and spatio-temporal constraints. 
As events are formally defined as entities that exist in time and not in space (cf. Section~\ref{sec:events}), constraints including spatial restrictions are expressed through the \objects participating in the component event.
For instance, a \fconcept{Component} event may be required to occur within a certain time-interval, e.g., the second week of June 2009. 
Depending on its objects, a \fconcept{Component} event may also happen in a certain spatial region.
For example, the flooding of a town should be composed of events that have objects associated to it, which have some certain range of longitude and latitude. 
Finally, events and the \objects bond to it may be qualified by a spatio-temporal quality like the progress of a flood that extents over time and space, starting with a high water level located in some area of a river and extending spatially over time into other areas. 
Any such constraints are formally expressed by one or multiple instances of the \dolconcept{F:EventCompositionConstraint}.
Thus, with the composition pattern, events may be arbitrarily temporally related to each other, i.e., they might be disjoint, overlapping, or otherwise ordered. 
In order to express such relative temporal relations between events, one can facilitate the provided means of DOLCE such as the formalization of Allen's Time Calculus\footnote{\url{http://wiki.loa-cnr.it/index.php/LoaWiki:Ontologies}}.

\subsection{Causality Pattern}
\label{sec:causalitypattern}

Causality is the 
philosophical problem investigating the existence 
of any special ``tie'' binding causes and effects together~\cite{ItkonenCausality1983}.
It can be questioned either as ``Why'' or ``How''~\cite{ItkonenCausality1983}.
An answer to this question is (or better said claims to be) a causal explanation.
What explains, is the cause and that what is explained is the effect~\cite{ItkonenCausality1983}.
Events are the most natural concept to serve for defining causal relations~\cite{QuintonObjectsAndEvents1979}.
In fact, causes and effects are two specific types of events~\cite{ItkonenCausality1983,LombardEvents1986}. 
A causal relationship is always justified by some (maybe implicit) underlying causal theory. 

We designed a causality pattern as depicted in Fig\-ure~\ref{fig:causality-pattern}(c). 
The pattern defines two \dolconcept{EventType}s called \fconcept{F:Cause} and \fconcept{F:Effect} which \dolconcept{classify} \dolconcept{Event}s. 
It further defines a \dolconcept{DUL:Description}, which is classified by a \dolconcept{F:Justification}.
By this, the pattern explicitly expresses the causal relationship between the cause and the effect under the justification of some theory.
A theory might be an opinion, a scientific law, or not further specified. 
For example, during a heavy storm, a power outage might occur caused by a snapped power pole. 
The \dolconcept{Justification} of this causal relationship is the laws of physics.

\subsection{Correlation Pattern}\label{sec:correlation}

A set of events is called correlated if they have a common cause.
However, there exists no causal relationship between the two events~\cite{ShipleyCausAndCorrelation2002}.
The common cause may originate from a single or a chain of multiple preceding cause-effect relationships. 
Correlation also differs from co-occurrence where two or more events just (randomly) happen at the same time and do not have a common cause.
Correlation is not of metaphysical interest as it is a property that can be derived from causality, i.e., the common cause.
In our ontology, we model correlation explicitly as in many cases the (correlating) effects of some common cause may be known, while the cause itself is not.
The correlation pattern depicted in Figure~\ref{fig:correlation}(d) defines the role \dolconcept{F:Correlate} to \dolconcept{classify} the events that are correlated.
The \dolconcept{Justification} role classifies some \dolconcept{Description}, which explains the correlation in terms of a (mathematical) law or some theory.

\subsection{Documentation Pattern}

Documentary evidence for an event may be given by arbitrary objects, e.g., some sensor data or media data, or by other events.
Formally, this relation is expressed by the documentation pattern depicted in Figure~\ref{fig:documentation-pattern}(e).
It defines the concept \dolconcept{F:DocumentedEvent} that classifies the documented event and the concept \dolconcept{F:Documenter} that classifies the documentary evidence for that event. 
This evidence can be expressed by any specialization of an \dolconcept{Object}, e.g., a digital photo taken with a cell phone during an incident, or a specialization of \dolconcept{Event}.
For example, digital media data like images and videos can be classified as \dolconcept{Documenter} and precisely described using the Core Ontology for Multimedia~\cite{ArndtEtAlCOMM2007}. 
Objects are documented via the events in which they participate (see participation pattern in Section~\ref{sec:participationpattern}).

\subsection{Interpretation Pattern}\label{sec:interpretation}

The perception of events as occurrences in the real world heavily depends on the context and point of view of the observer.
Such different, context-dependent event interpretations can be described formally by instantiating the different Event-Model-F patterns presented so far and binding them together with the interpretation pattern depicted in Figure~\ref{fig:interpretation-pattern}(f).
Each pattern models a single, specific interpretation of an event by associating \emph{participations}, \emph{mereological}, \emph{causal}, and \emph{correlative} relationships, as well as \emph{documentations} relevant in the context of a specific \emph{interpretation}. 
In the emergency use case, two emergency control officers might have differing interpretations of the power outage.
One might be convinced that the power outage is due to a snapped power pole, while the other might think of a more serious case of a damaged power plant.
Both consider the same event of a power outage, however, consider it from different points of view that involve other events and objects in different patterns.

Formally, the interpretation pattern shown in Figure~\ref{fig:interpretation-pattern}(f) defines a \fconcept{F:Interpretant} that is specialized from \dolconcept{EventType} and classifies the interpreted \dolconcept{Event}. 
The \fconcept{Interpretant} might be defined in some domain ontology and determines how an event is interpreted, e.g., as emergency incident in the case of the emergency control center or as news event described in a news paper.
Within each interpretation, we classify the \fconcept{F:RelevantSituation}s, namely the situations satisfying the participation, mereology, causality, correlation, and documentation.
These are defined as specializations of \dolconcept{RelevantSituation}.

\subsection{Non-functional Design Aspects}\label{sec:methdology}

For designing the Event-Model-F, we have aligned it with the foundational ontology DUL.
This choice of DUL as modeling basis specifically reflects the non-functional requirements (see Section~\ref{sec:nonfunc-requirements}).
Foundational ontologies provide a high-level, abstract vocabulary of concepts and relations that are likely to be used in current and future application domains. 
Thus, a precise alignment of concepts defined in the Event-Model-F with the high-level concepts of a foundational ontology provides a solid basis for future extensions 
(cf. requirement (a) on extensibility).
This alignment also includes the adoption and specialization of the formal semantics of the foundational ontology in our Event-Model-F ontology.
Thus, it supports for validating the more specific semantics of the concepts and relations defined in the Event-Model-F~(b).
For designing the Event-Model-F, we built upon the foundational ontology DOLCE that supports a pattern-oriented design approach~\cite{OberleEtAlOntologiesForLargeSoftwareSystems2006}.
We have designed the Event-Model-F as a set of patterns that structure the domain 
into smaller and better manageable ontology modules~(c).
The individual events and objects are reified by the different patterns of our ontology.
Splitting up the description of events and objects into different parts allows for reusing them among different applications, where only parts of the descriptions might be needed and combined with domain-specific knowledge~(d).
Finally, the separation of concerns (e) is supported by defining the structural knowledge of events and objects in the Event-Model-F and leaving all domain-specific aspects out of it. 
By this, the Event-Model-F is independent of any concrete domain that makes use of events and objects.

\section{Use of Event-Model-F}

\label{sec:use}

With the Event-Model-F, we can create and exchange sophisticated, formal descriptions of events.
We demonstrate the use of our ontology at the emergency response scenario described in Section~\ref{sec:scenario}. 
For describing an event, the different patterns defined in the Event-Model-F are combined, each providing a specific part of the event description.
To model the participation of citizens in an emergency incident like a flood one may use the participation pattern.
A flood may typically be composed of multiple events, which is modeled using the mereology pattern.
One can model the cause for the flood applying the causality pattern and possibly also using the correlation pattern to model correlating events.
As there might be different opinions about the cause of the flood, there can also be multiple instantiations of the causality pattern referring to the same events.
To manage these multiple instantiations of the causality pattern (or other pattern), the event interpretation pattern is used to form different nexuses of the pattern instances and providing different points of view onto the same event.
Thus, the interpretation pattern supports reusing parts of event descriptions on the level of ontology pattern.
In the following, we present two examples of applying our ontology patterns.

During a flood, a snapped power pole may cause a power outage.
This causal relationship between the two events is modeled using the causality pattern shown in the upper part of Figure~\ref{fig:example}(g).
It defines two \dolconcept{Event} individuals \individual{snapped-power-pole-1} and \individual{power-outage-1}, where the first one is classified as \dolconcept{Cause} and the latter as its \dolconcept{Effect}.
Some houses and citizens are involved in the \individual{power-outage-1} event, which is modeled using the participation pattern.
The lower part of Figure~\ref{fig:example}(g) shows an example, the individual \individual{person-1} living in the affected building \individual{house-1}. 
They are classified by the individuals \individual{citizen-1} and \individual{affected-bldg-1} in the participation pattern playing the roles of a \domainconcept{Citizen} and an \domainconcept{AffectedBuilding}.
Both roles are defined in a domain-specific emergency response ontology and reused here.
The \individual{power-outage-1} event described by this participation pattern is classified as such by an instance of the concept \dolconcept{DescribedEvent}.

In emergency response, the entities involved need to exchange event descriptions like the one above.
However, they use different systems with proprietary data models for describing events.
Thus, a formal model like the Event-Model-F may help to integrate them and to effectively communicate event descriptions.

\section{Existing Event Models}
\label{sec:relatedwork}

\begin{figure}[htb]
 \centering
\includegraphics[width=1\columnwidth]{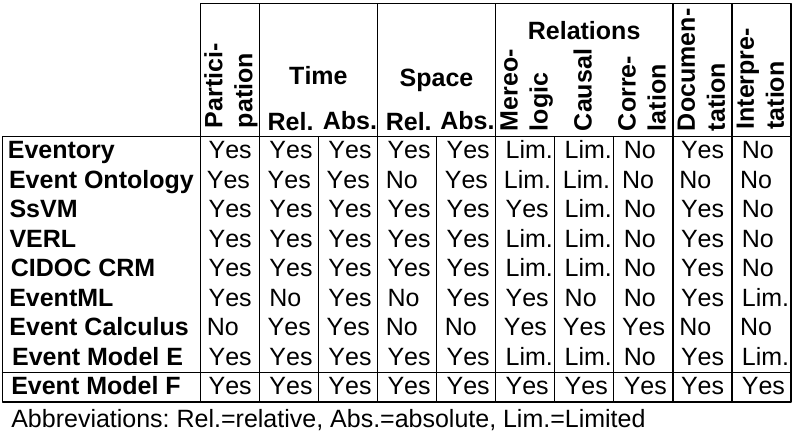}
 \caption{Comparison of Event Models} 
 \label{fig:eventmodelrequirementsmatrix}
\end{figure}

For designing the Event-Model-F, we analyzed existing event-based systems and event models with respect to the functional requirements. 
These models are motivated from different domains such as 
the Eventory~\cite{WangEtAl2007} system for journalism, 
the Event Ontology~\cite{RaimondAbdallahEventOntology2007} as part of a music ontology framework,
the ISO-standard of the International Committee for Documentation on a Conceptual Reference Model (CIDOC CRM)~\cite{DoerrCIDOC2007,SinclairEtAlCRMCore2006} for cultural heritage, 
the event markup language EventML~\cite{IPTCEventML2008} for news,
the event calculus~\cite{Mueller2008,Cervesato99aguided} for knowledge representation, 
the Semantic-syntactic Video Model (SsVM)~\cite{EkinEtAl2004} and 
Video Event Representation Language (VERL)~\cite{FrancoisEtAlVERL2005,NevatiaEtAlVERL2004} for video data,
and the event model E~\cite{ScherpEtAlEMMa2008,WestermannJainE2007} 
for event-based multimedia applications.
An overview of the analysis results and comparison to the features of our Event-Model-F along the functional requirements is shown in Figure~\ref{fig:eventmodelrequirementsmatrix}.

Our analysis shows that the existing event models almost fully support the participative, temporal, spatial, and documentary aspects.
However, existing event models substantially lack in supporting the structural aspect, i.e., mereological, causal, and correlation relationships, and representation of different interpretations of the same event.
Here, we find different variations of limitations or even no support by the existing event models.
With respect to mereological relationships, the existing event models typically provide support for simple part-of relationships such as with the \texttt{sub\_event} property of the Event Ontology. 
Only SsVM allows for describing more complex mereological relationships.
Similar, also the support for causal relationships is limited in the existing event models.
It is typically defined as a cause-effect relationship such as 
the ``resulted in'' property in CIDOC CRM.
Also no further axiomatization of causality is provided. 
The correlation relationship is only supported by a specific extension of the event calculus~\cite{Cervesato99aguided}.
None of the existing models provides for different interpretations of the same event.
Only E 
considers it as future work. 
Considering the functional requirements, we can state that the Event-Model-F supports all of them.

In addition, there is also much work in processing of low-level signals~\cite{MuhlEtAlDEBS2006}.
Here, an event is the (non-)existence of a signal. 
Relevant work are publish/subscribe systems and middleware solutions~\cite{OberleMiddleware2006}, complex event processing~\cite{EricssonBerndtssonREX2007} and event stream processing~\cite{Zdonik2003}, Semantic Web services~\cite{Vaculin2007}, and reactivity for the Semantic Web~\cite{May2005}.
This work focuses on technical events that happen \textit{within} computerized systems~\cite{MuhlEtAlDEBS2006}.
It cannot be used to model different event interpretations and event points of view as it is supported by the Event-Model-F.
With respect to the non-functional requirements, one can say that the existing event models 
do not follow a systematic development approach and do not define a formal semantics.
They also do not follow a pattern-oriented approach 
to structure the complex problem of an event model into smaller, reusable units.
Only the Event Ontology supports reuse of existing ontologies.
Extensibility and separation of concern are typically not in the scope of the existing event models.
As discussed in Section~\ref{sec:methdology}, our Event-Model-F fulfills all non-functional requirements.

\section{Conclusions} 

We designed a formal model of events based on the foundational ontology \DULextended.
By the use of this ontology and a pattern-oriented design approach, we are able to fulfill the non-functional requirements stated in Section~\ref{sec:requirements}.
For the functional requirements, we introduced specific ontology patterns based on the Descriptions and Situations model.
By this, we can represent arbitrary occurrences in the real world and formally model the different relations and interpretations of events.
The full support for the structural aspect as well as different event interpretations distinguishes the Event-Model-F very much from existing event models.
Separating the model into smaller patterns allows for better managing the complexity of events.
Due to its formal nature, our Event-Model-F allows for integration of different event-based systems and components.
In a future extension of our work, we plan to conduct reasoning on the instantiations of our patterns based on meta-knowledge. 
We also like to investigate the reusability of our Event-Model-F and its possibility to combine it with other core ontologies.

\textit{Acknowledgment.}
The first author kindly thanks Kai Wehmeier for his 
discussions about the philosophical basis of this work.
This research has been co-funded by the EU in FP6 in the X-Media project (026978) and FP7 in the WeKnowIt project (215453).

\balancecolumns


\begin{thebibliography}{10}

\bibitem{ArndtEtAlCOMM2007}
R.~Arndt, R.~Troncy, S.~Staab, L.~Hardman, and M.~Vacura.
\newblock {COMM}: Designing a well-founded multimedia ontology for the web.
\newblock In {\em ISWC}. Springer, 2007.

\bibitem{StanfordEncyclopediaEvents}
R.~Casati and A.~Varzi.
\newblock Events.
\newblock Stanford Encyclopedia of Philosophy, 2006.
\newblock http://plato.stanford.edu/entries/events.

\bibitem{Cervesato99aguided}
I.~Cervesato, M.~Franceschet, and A.~Montanari.
\newblock A guided tour through some extensions of the event calculus.
\newblock {\em Computational Intelligence}, 16:200--0, 1999.

\bibitem{DoerrCIDOC2007}
M.~Doerr, C.-E. Ore, and S.~Stead.
\newblock The {CIDOC} conceptual reference model: a new standard for knowledge
  sharing.
\newblock In {\em Conceptual modeling}. Australian Computer Society, Inc.,
  2007.

\bibitem{EkinEtAl2004}
A.~Ekin, A.~M. Tekalp, and R.~Mehrotra.
\newblock Integrated semantic-syntactic video modeling for search and browsing.
\newblock {\em IEEE Transactions on Multimedia}, 6(6), 2004.

\bibitem{EricssonBerndtssonREX2007}
A.~Ericsson and M.~Berndtsson.
\newblock Rex, the rule and event explorer.
\newblock In H.-A. Jacobsen, G.~M{\"u}hl, and M.~A. Jaeger, editors, {\em
  DEBS}. ACM, 2007.

\bibitem{FrancoisEtAlVERL2005}
A.~R.~J. Francois, R.~Nevatia, J.~Hobbs, and R.~C. Bolles.
\newblock {VERL}: An ontology framework for representing and annotating video
  events.
\newblock {\em IEEE MultiMedia}, 12(4), 2005.

\bibitem{FranzStaabArndtXCOSIM2007}
T.~Franz, S.~Staab, and R.~Arndt.
\newblock The {X-COSIM} integration framework for a seamless semantic desktop.
\newblock In {\em Knowledge capture}. ACM, 2007.

\bibitem{GangemiDnS2008}
A.~Gangemi.
\newblock Norms and plans as unification criteria for social collectives.
\newblock {\em Autonomous Agents and Multi-Agent Systems}, 17(1), 2008.

\bibitem{GangemiEtAlDOLCE2002}
A.~Gangemi, N.~Guarino, C.~Masolo, A.~Oltramari, and L.~Schneider.
\newblock {Sweetening Ontologies with DOLCE}.
\newblock In {\em EKAW}. Springer, 2002.

\bibitem{IPTCEventML2008}
{IPTC}.
\newblock {EventML}, 2008.
\newblock http://iptc.org/.

\bibitem{ItkonenCausality1983}
E.~Itkonen.
\newblock {\em Causality in Linguistic Theory}.
\newblock Indiana Univ. Press, 1983.

\bibitem{LombardEvents1986}
L.~Lombard.
\newblock {\em Events: A metaphysical study}.
\newblock Routledge \& Kegan Paul, 1986.

\bibitem{DOLCE2003}
C.~Masolo, S.~Borgo, A.~Gangemi, N.~Guarino, and A.~Oltramari.
\newblock {WonderWeb} deliverable {D18} ontology library; {IST WonderWeb}
  project, 2003.

\bibitem{May2005}
W.~May, J.~J. Alferes, and R.~Amador.
\newblock An ontology- and resources-based approach to evolution and reactivity
  in the semantic web.
\newblock In {\em OTM}. Springer, 2005.

\bibitem{Mueller2008}
E.~T. Mueller.
\newblock {\em Handbook of Knowledge Representation}, chapter Event Calculus.
\newblock Elsevier, 2008.

\bibitem{MuhlEtAlDEBS2006}
G.~M\"{u}hl, L.~Fiege, and P.~Pietzuch.
\newblock {\em Distributed Event-Based Systems}.
\newblock Springer, 2006.

\bibitem{NevatiaEtAlVERL2004}
R.~Nevatia, J.~Hobbs, and B.~Bolles.
\newblock An ontology for video event representation.
\newblock In {\em Computer Vision and Pattern Recognition}. IEEE, 2004.

\bibitem{OberleMiddleware2006}
D.~Oberle.
\newblock {\em Semantic Management of Middleware}.
\newblock Springer, 2006.

\bibitem{OberleEtAlOntologiesForLargeSoftwareSystems2006}
D.~Oberle and {et al.}
\newblock Towards ontologies for formalizing modularization and communication
  in large software systems.
\newblock {\em Appl. Ontol.}, 1(2):163--202, 2006.

\bibitem{OberleEtAlSWIntO2007}
D.~Oberle and {et al.}
\newblock {DOLCE} ergo {SUMO}: On foundational \& domain models in the
  {SmartWeb Integrated Ontology}.
\newblock {\em Web Semant.}, 5(3), 2007.

\bibitem{QuintonObjectsAndEvents1979}
A.~Quinton.
\newblock Objects and events.
\newblock {\em Mind}, 88(350), 1979.

\bibitem{RaimondAbdallahEventOntology2007}
Y.~Raimond and S.~Abdallah.
\newblock The event ontology, 2007.
\newblock http://motools.sf.net/event.

\bibitem{ScherpEtAlEMMa2008}
A.~{Scherp}, S.~{Agaram}, and R.~{Jain}.
\newblock {Event-centric media management}.
\newblock In {\em SPIE}, 2008.

\bibitem{ShipleyCausAndCorrelation2002}
B.~Shipley.
\newblock {\em Cause and Correlation in Biology}.
\newblock {Cambridge Univ. Press}, 2002.

\bibitem{SinclairEtAlCRMCore2006}
P.~Sinclair, M.~Addis, F.~Choi, and {et al.}
\newblock The use of {CRM} core in multimedia annotation.
\newblock In {\em Semantic Web Annotations for Multimedia}, 2006.

\bibitem{Vaculin2007}
R.~Vaculin and K.~Sycara.
\newblock Specifying and monitoring composite events for semantic web services.
\newblock In {\em ECOWS}. IEEE, 2007.

\bibitem{WangEtAl2007}
X.~Wang, S.~Mamadgi, A.~Thekdi, A.~Kelliher, and H.~Sundaram.
\newblock Eventory -- an event based media repository.
\newblock In {\em Semantic Computing}. IEEE, 2007.

\bibitem{WestermannJainE2007}
U.~Westermann and R.~Jain.
\newblock Toward a common event model for multimedia applications.
\newblock {\em IEEE MultiMedia}, 14(1), 2007.

\bibitem{Zdonik2003}
S.~B. Zdonik, M.~Stonebraker, M.~Cherniack, and {et al.}
\newblock The {A}urora and {M}edusa projects.
\newblock {\em IEEE Data Eng.}, 26(1), 2003.
\end{thebibliography}
\end{document}